\documentclass{article}
\usepackage{ijcai26}

\usepackage{times}
\usepackage{soul}
\usepackage{url}
\usepackage[hidelinks]{hyperref}
\usepackage[utf8]{inputenc}
\usepackage[small]{caption}
\usepackage{graphicx}
\usepackage{amsmath}
\usepackage{amsthm}
\usepackage{booktabs}
\usepackage{algorithm}
\usepackage{algorithmic}
\usepackage[switch]{lineno}
\usepackage{multirow}
\usepackage{amssymb}
\usepackage{booktabs}
\usepackage{tabularx, booktabs}
\usepackage{graphicx}
\usepackage{tabularx}
\usepackage{subcaption}
\title{LLMs for High-Frequency Decision-Making: Normalized Action Reward-Guided Consistency Policy Optimization}

\author{
Yang Zhao$^1$\and
Zihao Li$^1$\and
Zhiyu Jiang$^1$\and
Dandan Ma$^1$\and
Ganchao Liu$^1$\and
Wenzhe Zhao$^1$\\
\affiliations
$^1$School of Artificial Intelligence, OPtics and ElectroNics (iOPEN), Northwestern Polytechnical University, Xi’an 710072, China\\
\emails
izhaoyang@nwpu.edu.cn,
lzhiopen@mail.nwpu.edu.cn
}

\begin{document}

\maketitle

\begin{abstract}
    While Large Language Models (LLMs) form the cornerstone of sequential decision-making agent development, they have inherent limitations in high-frequency decision tasks. Existing research mainly focuses on discrete embodied decision scenarios with low-frequency and significant semantic differences in state space (e.g., household planning). These methods suffer from limited performance in high-frequency decision-making tasks, since high-precision numerical state information in such tasks undergoes frequent updates with minimal fluctuations, and exhibiting policy misalignment between the learned sub-tasks and composite tasks. To address these issues, this paper proposes Normalized Action Reward guided Consistency Policy Optimization (NAR-CP). 1) Our method first acquires predefined dense rewards from environmental feedback of candidate actions via reward functions, then completes reward shaping through normalization, and theoretically verifies action reward normalization does not impair optimal policy. 2) To reduce policy misalignment in composite tasks, we use LLMs to infer sub-observation candidate actions and generate joint policies, with consistency loss ensuring precise alignment between global semantic policies and sub-semantic policies. Experiments on UAV pursuit, a typical high-frequency task, show our method delivers superior performance on independent and composite tasks with excellent generalization to unseen tasks.

\end{abstract}

\section{Introduction}

    Large language models (LLMs), empowered by their strong semantic understanding ability and vast prior knowledge, have emerged as the core driving force for the development of sequential decision-making agents~\cite{huang2022inner}~\cite{brohan2023can}~\cite{carta2023grounding}~\cite{chung2024scaling}. They exhibit enormous potential to break through the sample efficiency bottleneck of traditional reinforcement learning (RL) in fields such as embodied interaction and task planning~\cite{chevalier2018babyai}~\cite{szot2023large}~\cite{carta2023grounding}~\cite{tan2024true}. As decision-making agents, LLMs are required to achieve accurate alignment with environmental dynamics and optimize decision-making policy through trial-and-error learning, thereby robustly complete complex tasks.

    The current advanced methods have effectively addressed the problems of environment alignment and action optimization in discrete scenarios through techniques such as token normalization~\cite{tan2024true} and token-level credit allocation~\cite{wen2024reinforcing}, significantly improving decision-making efficiency and sample utilization rate. However, their core designs are all tailored for discrete embodied decision-making scenarios with short-term (fewer decision steps) and distinct semantic differences in the state space (e.g., household planning, kitchen collaboration~\cite{tan2024true}~\cite{wen2024reinforcing}). By contrast, in continuous decision-making scenarios, tasks exhibit core characteristics such as high-frequency temporal dynamic evolution, observations containing high-precision numerical text, and the need for composite task decision-making. Existing methods fail to adapt to the numerical-semantic hybrid observation characteristics of high-frequency continuous decision-making and no attention was paid to the policy bias of the large language model in the composite task.
    
    Consequently, they have significant limitations in two core directions: 

\begin{enumerate}
    \item \textbf{Policy convergence problem in high-frequency decision making environments:} In high-frequency sequential decision-making tasks, observations consist of high-precision numerical text and environmental descriptions. Minimal numerical variations between consecutive mixed observations lead to high semantic overlap, making it difficult for the model to discern environmental dynamics from raw textual representations, and the weak and sparse reward feedback is difficult to provide effective supervision for action value. Consequently, the model tends to misclassify similar states into identical semantic categories, impairing the establishment of accurate state-action value mapping and ultimately leading to difficulties in training convergence and environmental alignment.
    \item \textbf{Policy bias in LLM composite task:} The autoregressive generation property inherent in large language models imposes strong input dependence on their policy learning, thereby impeding the transfer of effective knowledge acquired in  sub-tasks to composite task scenarios. When composite observations are input, the generated composite policy differs significantly from the product of sub-task policy ($\pi(a_1 a_2 \mid s_1 s_2) \neq \pi(a_1 \mid s_1) \times \pi(a_2 \mid s_2)$).
\end{enumerate}
    To address policy convergence issues caused by sparse rewards in high-frequency environments, traditional RL has proposed several countermeasures. Reward shaping introduces intermediate feedback signals to bridge reward gaps~\cite{ng1999policy,grzes2017reward,hu2020learning}, curriculum learning reduces exploration difficulty by progressively increasing task complexity~\cite{bengio2009curriculum,narvekar2020curriculum}, and model-based RL improves sample efficiency by learning environment dynamics~\cite{sutton1996model,clavera2018model,m2023model}. While these methods aim to stabilize training and accelerate state--action correlation learning, they face intrinsic limitations when applied to LLM-driven high-frequency decision-making. Unlike conventional RL operating on structured numerical representations, LLMs encode precise values into text, where subtle reward differences are difficult to distinguish, leading to ambiguous state--action equivalence and impaired convergence.

    For policy bias in LLM-based composite tasks, existing approaches exhibit clear limitations. Chain-of-Thought prompting enhances coordination reasoning~\cite{wei2022chain,feng2023towards}, but its high inference latency hinders real-time sequential decision-making. Alternatively, observation decoupling and policy concatenation simplify training~\cite{xu2023rewoo}, yet disrupt semantic coherence, causing policy inconsistencies when composite tasks are encountered with unified observations.
 
    In this paper, we enhance the discrimination of rewards through normalized action reward  to optimize strategy learning, and theoretically discuss that our method will not influence the optimal strategy. We propose a policy consistency learning architecture with observation decoupling, which retains the advantages of decoupled training while reducing the deviation between the composite task strategy and the sub-task joint strategy. Our main contributions are as follows:

\begin{itemize}
\item Extend LLM-based agents to high-frequency decision-making scenarios, propose the normalized action reward has alleviated the problem of policy learning failure caused by low discrimination rewards.
    
\item Propose an observation-decoupled consistency policy learning method that retains the advantage of decoupled training and enables consistent policy performance of LLMs on sub-tasks and composite task.
    
\item Taking UAV tracking as the high-frequency decision-making research subject, we design pertinency simulation experiments to verify the proposed method's task performance and generalization ability in this scenario.
\end{itemize}

\section{Preliminaries}

\paragraph{Dense Reward shaping in RL.}

The sequential decision-making problem in traditional reinforcement learning can be formalized within the {\em Markov Decision Process} (MDP)~\cite{puterman1990markov} framework $\mathcal{M} = (\mathcal{S}, \mathcal{A}, T, R, \gamma)$, where $\mathcal{S}$ denotes the state space, $\mathcal{A}$ denotes the action space, $T: \mathcal{S} \times \mathcal{A} \to \Delta(\mathcal{S})$ represents the state transition probability function, $R: \mathcal{S} \times \mathcal{A} \times \mathcal{S} \to \mathcal{R}$ is the reward function, and $\gamma \in [0, 1)$ is the discount factor. The primary objective is to learn an optimal policy $\pi^*: \mathcal{S} \to \Delta(\mathcal{A})$ that maximizes the expected cumulative discounted reward over a trajectory $\tau = (s_0, a_0, \dots, s_T, a_T)$.

However, current methods often adopt a terminal reward form, providing non-zero feedback only at the terminal time step. This sparse feedback characteristic easily leads to slow convergence during policy training. To address this issue, introduce a dense reward shaping mechanism~\cite{grzes2017reward}, defining a new MDP process $M_D = (\mathcal{S}, \mathcal{A}, T, R_D, \gamma)$. Here, the dense reward function is defined as $R_D = R + F$, where the additive term $F = \gamma \phi(s_{t+1}) - \phi(s_t)$~\cite{ng1999policy}. The function $\phi$ serves as a potential function that quantifies the distance between the current state and the goal state.

Under this mechanism, the optimal action-value function satisfies the Bellman optimality equation:
\begin{equation}
\begin{aligned}
    Q_D^*(s, a) = {E}_{s' \sim P(s, a)} [&R(s, a, s') 
    + \gamma \phi(s') - \phi(s)+\\&\gamma \max_{a' \in \mathcal{A}} (Q^*(s', a') - \phi(s')) ].
\end{aligned} 
\end{equation}

At optimality, the optimal policy of \(M_D\) is given by :
\begin{equation}
\begin{aligned}
    \pi_{M_D}^*(s) &\in \arg\max_{a \in \mathcal{A}} Q_{M_D}^*(s, a) \\&= \arg\max_{a \in \mathcal{A}} Q^*(s, a) - \phi(s).
\end{aligned}
\end{equation}
\(\phi(s)\) is solely a function of state \(s\) and action-independent, the correction magnitude remains uniform across all actions; consequently, maximizing   $ Q^*(s, a) - \phi(s) $   is mathematically equivalent to maximizing the original   $ \mathrm{MDP} $     $ Q^*(s, a) $  , ensuring that the optimal policy   $ \pi^* $   coincides with   $ \pi_{M_D}^* $   and that the dense reward mechanism introduces no policy bias.
    
\begin{figure*}
    \centering
    \includegraphics[width=0.8\linewidth]{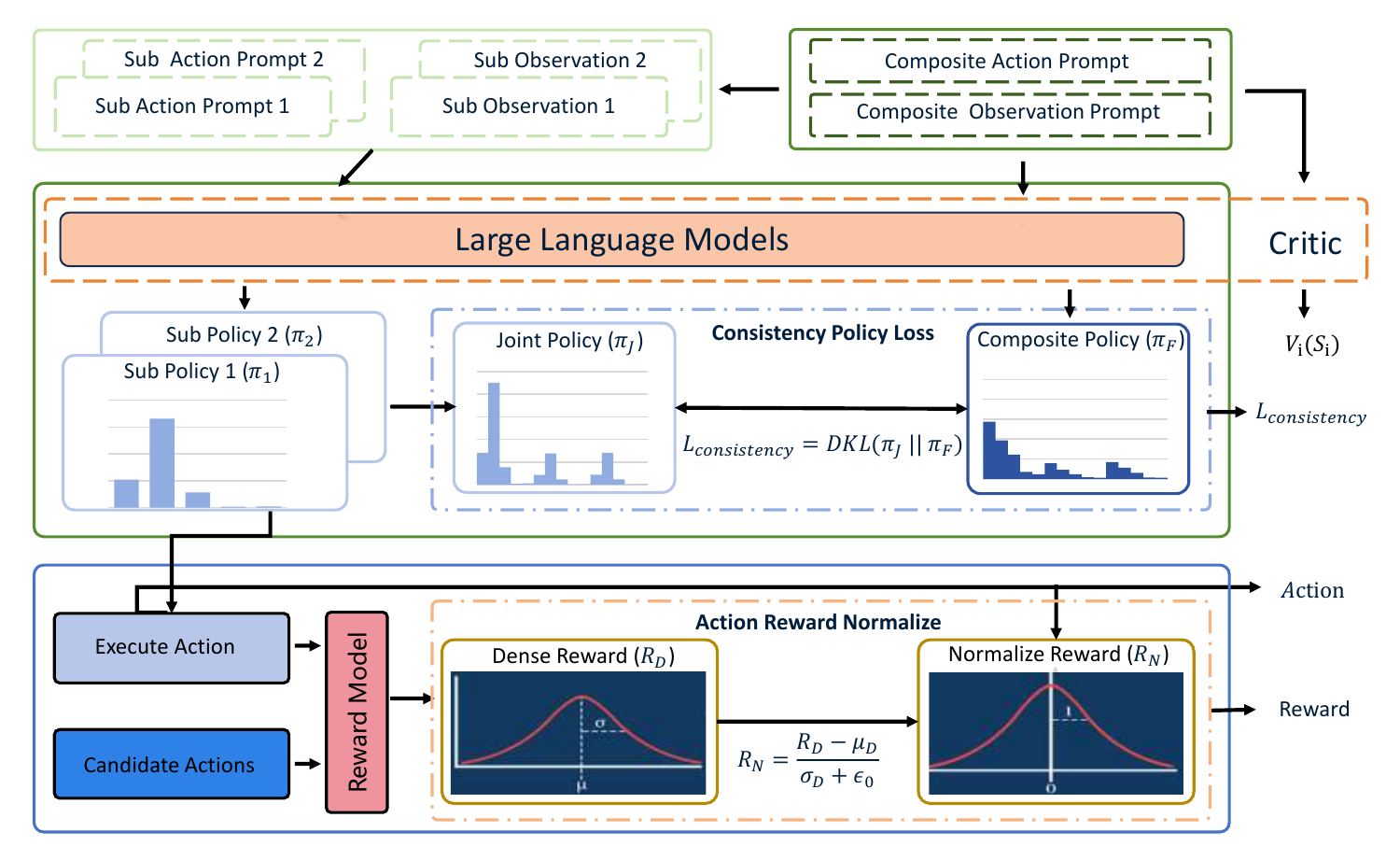}
    \caption{Method Overview. (a) Reduce the deviation between the sub-tasks joint policy $\pi_J$ and the composite policy $\pi_F$ through the consistency loss. (b) By applying Action Reward Normalization to the original dense reward, we dynamically amplify the reward variance and restore the effectiveness of gradient signals. }
    \label{fig:placeholder}
\end{figure*}

\section{NAR-CP}

\subsection{Dense Reward to Language-augmented RL}
    For large language model reinforcement learning scenarios dependent on precise numerical values, rewards are no longer derived from text state but computed directly from the environment's original numerical state to ensure accuracy. 

    To this end, we formalize this problem as a numerically enhanced Markov decision process $M_{\text{num-LLM}} = (S_{\text{num}}, S_{\text{text}}, A_{\text{text}}, T, R_{\text{dense}}, \gamma, \Phi)$. $S_{\text{num}} \subseteq \mathbb{R}^n$ denotes the environment's original numerical state space containing precise observations required for continuous decision-making such as distance measurements and precision metrics; $T: S_{\text{num}} \times A_{\text{text}} \to S'_{\text{num}}$ represents the environmental state transition function that describes the objective mapping of numerical states and actions to subsequent numerical states; $R_D: S_{\text{num}} \times A_{\text{text}} \to \mathbb{R}$ is the dense reward function whose values are calculated directly from the original numerical states and actions;  $\Phi: S_{\text{num}} \to \mathbb{R}^d$ constitutes the nonlinear numerical-to-text-to-vector conversion function that transforms numerical states into semantic vectors.

    In high-frequency continuous decision-making environments, the precision of rewards depends on subtle variations in numerical states, resulting in extremely small reward differences for different actions under identical numerical states—the difference concentrates in the $10^{-3} \sim 10^{-2}$ magnitude :
\begin{equation}
    D\big(R_D(s, \mathcal{A}_{\text{text}})\big) = \sup_{a,a' \in \mathcal{A}_{\text{text}}} \big| R_D(s, a) - R_D(s, a') \big|.
\end{equation}
    $D(X)$ denotes the diameter of the set $X$; $\sup$ represents the supremum operator (least upper bound); for all numerical states $\forall s \in \mathcal{S}_{\text{num}}$, the reward diameter must satisfy $D(R_D(s, \mathcal{A}_{\text{text}})) \leq \kappa$ (with $\kappa \in [10^{-3}, 10^{-2}]$). Together, they characterize the upper bound of reward differences across all possible action pairs in $\mathcal{A}_{\text{text}}$. 

    Considering the reward distribution centered at the reward mean $\mu(s) = \mathbb{E}_{a \sim \pi}[R_D(s,a)]$, all possible action rewards lie within the interval $[\mu(s)-D(X)/2, \mu(s)+D(X)/2]$ , strictly constraining the reward dispersion. Consequently, for any state $s \in \mathcal{S}_{\text{num}}$, the theoretical upper bound of reward variance relates deterministically to the reward diameter as:
\begin{equation}
    \text{Var}[R_D(s,\mathcal{A}_{\text{text}})] \leq \frac{D(X)^2}{4}, \quad \forall s \in \mathcal{S}_{\text{num}}.
\end{equation}

    The variance of the original gradient estimate can be expressed as:
\begin{equation}
    \text{Var}({\nabla} J(\theta)) = \frac{1}{n} \cdot \text{Var}(R_D) \cdot \mathbb{E}\left[\|\nabla_{\theta} \log \pi_{\theta}(a_{\text{text},i}|s_{\text{text},i})\|^2\right].
\end{equation}
    $\mathbb{E}\left[\|\nabla_{\theta} \log \pi_{\theta}(a_{\text{text},i}|s_{\text{text},i})\|^2\right]$ represents the mathematical expectation of the squared log-likelihood gradient with respect to the model parameters (denoted as $\mathbb{E}[\cdot]$)~\cite{wei2025redit}, which is a non-zero constant independent of the reward distribution.

    The gradient norm relates to the gradient variance as:
\begin{equation}
    \|\nabla J(\theta)\| \propto \sqrt{\text{Var}(\widehat{\nabla} J(\theta))}.
\end{equation}    
    when $\|\nabla J(\theta)\| < 10^{-2}$~\cite{wei2025redit}, significant gradient vanishing occurs, rendering model updates ineffective.
    
   The transformation $\Phi: S_{\text{num}} \to \mathbb{R}^d$ maps numerical values similar to highly analogous semantic representations, whose induced overlap reduces the distinguishability of input features and impairs the model's detection of subtle numerical variations. Even with reward signals accurately reflecting action values, such weak gradients fail to enable effective optimization, leading to ultimate training stagnation.   

\subsection{Normalized Action Reward}
    In this section, we introduce an action-reward normalization  mechanism. By applying Z-score normalization to the original dense reward , we dynamically amplify the reward variance and restore the effectiveness of gradient signals. Meanwhile, we theoretically guarantee that the optimal policy derived from the normalized reward is consistent with that of the original reward.
\paragraph{Normalize.} The original dense reward function is defined as $R_D: \mathcal{S}_{\text{num}} \times \mathcal{A}_{\text{text}} \mapsto \mathbb{R}$, which maps the environment's original numerical state $s_{\text{num}}$ and the candidate textual action $a_{\text{text}}$ to the corresponding dense reward value.  

    In timestep $t$, given the numerical state $s_{\text{num},t}$ observed in the environment, a set of candidate actions is provided $\mathcal{A}_t = \{a_i\}_{i=1}^{|\mathcal{A}_t|}$. This set encompasses all semantically valid and executable actions available under the current state.

    Based on this candidate set, we model the distribution of the original dense reward function $R_D$. Specifically, we first compute the first and second-order statistics: the mean $\mu_{D,t}$ and the standard deviation $\sigma_{D,t}$, as defined by the following equations:
\begin{equation}
\left\{
\begin{aligned}
    \mu_{D,t} &= \frac{1}{|\mathcal{A}_t|} \sum_{a_i \in \mathcal{A}_t} R_D(s_{\text{num},t}, a_i). \\
    \sigma_{D,t} &= \sqrt{\frac{1}{|\mathcal{A}_t|} \sum_{a_i \in \mathcal{A}_t} \left( R_D(s_{\text{num},t}, a_i) - \mu_{D,t} \right)^2 }.
\end{aligned}
\right.
\end{equation}
    $|\mathcal{A}_t|$ denotes the cardinality of the candidate action set, which serves as the batch size for the normalization calculation. These two statistics collectively constitute a dynamic representation of the reward distribution at the current timestep.

    We apply Z-score Normalization to transform the raw reward $R_D$. The normalized dense reward function $R_N$ is formally defined as:
\begin{equation}
    R_N(s_{\text{num},t}, a_i) = \frac{R_D(s_{\text{num},t}, a_i) - \mu_{D,t}}{\sigma_{D,t} + \epsilon_0},
\end{equation}
where $\epsilon_0$ is a small constant ($10^{-5}$) introduced to ensure numerical stability and prevent division by zero.

\paragraph{Gradient Norm.}After Z-score normalization, the original dense rewards at a single timestep are mapped to a unit variance distribution, explicitly amplifying the variance magnitude. The policy gradient variance is then expressed as:
\begin{equation}
    \text{Var}(\hat{\nabla} J(\theta)) = \frac{1}{|\mathcal{A}_t|} \cdot \text{Var}(R_N) \cdot \mathbb{E}[ \|\nabla_\theta \log \pi_\theta(a_i\mid s_{\text{text},t}) \|^2 ],
\end{equation}
where $|\mathcal{A}_t|$ denotes the number of candidate actions at the current timestep, and $\text{Var}(R_N) = 1$ holds by definition of the normalization. 

This operation reconstructs the gradient scale, driving the gradient norm $\|\nabla J(\theta)\| $ beyond the effective update threshold. Consequently, it ensures robustness and convergence of the policy optimization process even when the original reward variance is constrained.

\paragraph{Optimal Policy.}We defined a potential function based on the z-score normalization, which depends solely on the numerical state $s_{\text{num}, t}$ at a single time step $t$:
\begin{equation}
    \phi(s_{\text{num},t}) = \frac{\mu_{D,t}}{(\sigma_{D,t} + \epsilon_0)(1-\gamma)}.
\end{equation}

    $\mu_{D,t}$ and $\sigma_{D,t}$ are the reward statistics of candidate actions exclusive to $s_{\text{num},t}$, so this potential function is independent of actions; $\gamma$ is the discount factor.

    Based on Eq. (8) and the negligible state difference between adjacent time steps, the normalized reward can be reformulated as a new reward mechanism:
\begin{equation}
\begin{aligned}   
    R_N(s_{\text{num},t}, a_i) &= \frac{1}{\sigma_{D,t} + \epsilon_0} R_D(s_{\text{num},t}, a_i) \\&+ \gamma \phi(s'_{\text{num},t}) - \phi(s_{\text{num},t}).
\end{aligned}
\end{equation}    

    Substituting the  normalization reward into the Bellman optimality equation, the corresponding optimal action-value function can be derived:

\begin{equation}
\begin{aligned}
     Q_N^*(s_{\text{num}}, a) = &\mathbb{E}_{s'_{\text{num}} \sim P(  s_{\text{num}}, a)} 
    [
        \frac{1}{\sigma_{D} + \epsilon_0} R_D \\&+ \gamma \phi(s'_{\text{num}}) - \phi(s_{\text{num}}) + \gamma \max_{a' \in \mathcal{A}} (\\&\frac{1}{\sigma_{D} + \epsilon_0}Q_D^*(s'_{\text{num}}, a') - \phi(s'_{\text{num}})) 
    ].
\end{aligned}      
\end{equation}  
    $\mathbb{E}_{s'_{\text{num}} \sim P(s_{\text{num}},a)}$ denotes the expectation over the next numerical state $s'_{\text{num}}$ following the state transition distribution $P(s_{\text{num}},a)$; $R_d$ is the shorthand for $R_D(s_{\text{num}}, a)$ for brevity, and $\frac{1}{\sigma_{d,t} + \epsilon_0} > 0$ which is a positive scaling coefficient.

    The optimal policy under the normalization action reward can be expressed as:
    
\begin{equation}
\begin{aligned}
      \pi_{M_N}^*(s)  &\in \arg\max_{a \in \mathcal{A}} Q_N^*(s_{\text{num}}, a) \\&= \frac{1}{\sigma_{D} + \epsilon_0}\arg\max_{a \in \mathcal{A}} Q_D^*(s_{\text{num}}, a) - \phi(s_{\text{num}})\\&= \arg\max_{a \in \mathcal{A}} Q_D^*(s_{\text{num}}, a).
\end{aligned}      
\end{equation} 

    The positive scaling by $\frac{1}{\sigma_{D} + \epsilon_0}$ and the translation by the potential function $\phi(s_{\text{num}})$ do not alter the optimal values of the action-value function. Consequently, the optimal policy $\pi_{M_N}^*(s)$ under normalized rewards and the optimal policy $\pi_{M_D}^*(s)$ under original dense rewards exhibit no bias.

    Ultimately, through normalization action reward, we effectively enhance the gradient signal fidelity while preserving the optimality of the policy.
   
\subsection{Decoupled Consistency Policy}
    The decision-making process of language agents is modeled as a Language-Augmented Markov Decision Process (MDP)~\cite{carta2023grounding}. To bridge token-level language modeling with action-level policy optimization, the probability of a language action $P(a | s)$ is factorized into the product of its constituent tokens' probabilities~\cite{tan2024true}:
\begin{equation}
    P(a \mid s) = \prod_{i=1}^{|a|} P(w_i \mid s, w_{1:i-1}),
\end{equation}
    
    For complex tasks composed of independent sub-tasks, each sub-task possesses a distinct action space. The core requirement is to ensure consistency between the joint policy of learned sub-tasks and the composite task policy. Specifically, the composite policy under composite observations should align with the joint policy of sub-policies under decoupled observations, thereby enabling knowledge reuse to improve learning efficiency and reliability.

    The optimal policy for each independent sub-task $\pi_i$  is formulated as:
\begin{equation}
    \pi_i(a_i \mid s_i) = \frac{\exp\left(\log P_i(a_i \mid s_i) / L(a_i)\right)}{\sum_{a_i' \in \mathcal{A}_i} \exp\left(\log P_i(a_i' \mid s_i) / L(a_i')\right)}.
\end{equation}
    $a_i \in \mathcal{A}_i$, $s_i\in \mathcal{S}_i$, $L(a_i)$ respectively corresponding to the tasks $i$ action, state, and action normalization coefficient.

    Because the independence of sub-tasks, the joint action space is the Cartesian product $\{A_i\}=\prod_{i=1} \mathcal{A}_i$ composed of the action set $\{a_i\}$. The joint policy is defined as the product of individual policies:
\begin{equation}
    \pi_\text{Joint}(\{a_i\} \mid \{s_i\}) = \prod_{i=1} \pi_i(a_i \mid s_i),
\end{equation}
    where $\{a_i\}$ and $\{s_i\}$ represent the collections of actions and states.

    Our objective is to ensure that the composite task policy $\pi_\text{Full}(a \mid s)$ (with $a \in \{A_i\}$ and composite state $s$) remains consistent with $\pi_\text{Joint}(\{a_i\} \mid \{s_i\})$.

    To balance the preservation of policy consistency with reduced computational demands, we employ a Top-\(K\) mothod, select \(k\) actions with the highest probabilities from the joint action space $\{A_i\}$ of the -sub-task joint policy \(\pi_\text{Joint}(\{a_i\} | \{s_i\})\). These selected actions constitute the Top-\(K\) action set \(\mathcal{A}_{\text{top-k}}\ \subseteq \{A_i\}\) and \(|\mathcal{A}_{\text{top-k}}| = k\).

    We denote $ S=\{s_i\} $ and re-normalize the probabilities of actions within this subset to obtain the Top-\(K\) policy distribution \(\pi_{\text{J-topk}}\):
\begin{equation}
    \pi_{\text{J-top-k}}(a \mid s) = \frac{\pi_\text{Joint}(a \mid s)}{\sum_{a' \in \mathcal{A}_{\text{top-k}}} \pi_\text{Joint}(a' \mid s)}.
\end{equation}

    Compute the probability distribution of the composite task policy over the $\mathcal{A}_{\text{topk}}$, denoted as \(\pi_{\text{F-topk}}\):
\begin{equation}\begin{split}
    &\pi_{\text{Full-top-k}}(a \mid s) \\&= \frac{\exp\left(\log P(a \mid s) / L(a)\right)}{\sum_{a' \in \mathcal{A}_{\text{top-k}}} \exp\left(\log P(a' \mid s) / L(a')\right)}.
\end{split}
\end{equation}

    To minimize the distributional discrepancy between $\pi_{\text{J-topk}}$ and $\pi_{\text{F-topk}}$ within the Top-$K$ $\mathcal{A}_{\text{topk}}\ $ subset and ensure decision consistency, we employ the KL divergence to construct the consistency loss $\mathcal{L}_{\text{con}}$, which measures the overall fitness of the two distributions:
\begin{equation}
    \mathcal{L}_{\text{consistency}} = D_{\text{KL}} \left( \pi_{\text{J-topk}} \, \middle\| \, \pi_{\text{F-topk}} \right).
\end{equation}

\begin{figure}
    \centering
    \includegraphics[width=0.9\linewidth]{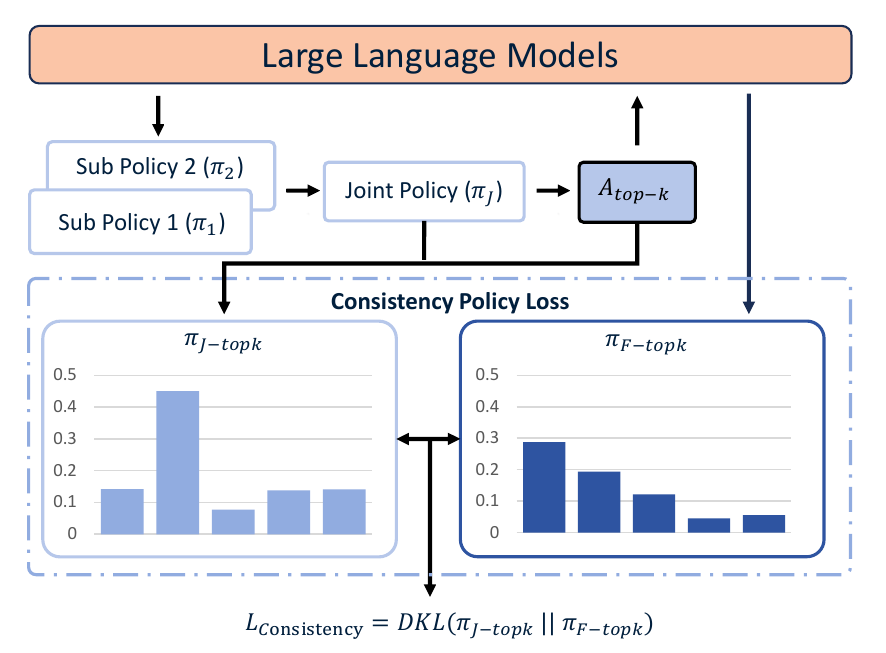}
    \caption{Details Of The  Consistency Policy. $\mathcal{A}_{\text{top-k}}$ represents the action with the highest probability extracted from the joint policy $\pi_J$ .\(\pi_{\text{Full-topk}}\) obtained by inputting the combination of $\mathcal{A}_{\text{top-k}}$ and composite observation prompt into the LLM. Finally employ the KL divergence to construct the consistency loss $\mathcal{L}_{\text{consistency}}$}
    \label{fig:placeholder2}
\end{figure}

    $D_{\text{KL}}(\cdot \, \| \, \cdot)$ denotes the KL divergence calculated over the Top-$K$ action set $\mathcal{A}_{\text{top-k}}$. By aggregating the probability contributions of the $k$ core actions, this term characterizes the overall distributional discrepancy and enforces alignment at the policy ensemble level.

    We formulate the overall policy loss function by combining the PPO loss, entropy regularization, and the consistency term. The unified expression is defined as:
\begin{equation}
\begin{aligned}  
    \mathcal{L}_{\text{policy}} = &-\mathbb{E} \left[ \min \left( \rho A, \text{clip}(\rho, 1-\varepsilon, 1+\varepsilon) A \right) \right] \\& - \alpha_{\text{e}} \mathcal{L}_{\text{entropy}} + \alpha_{\text{c}} \mathcal{L}_{\text{consistency}},
\end{aligned}  
\end{equation}
    where $\rho$ denotes the importance sampling ratio, calculated as :
\begin{equation*}
    \rho = \frac{ \pi_{J}(\{a_i\} \mid \{s_i\}) }{ \pi_{J,\text{old}}(\{a_i\} \mid \{s_i\}) }.
\end{equation*}

    $A$ represents the advantage function with  Generalized Advantage
Estimation; $\varepsilon$ is the PPO clipping range (typically $0.2$); $\mathcal{L}_{\text{entropy}}$ is the entropy loss with coefficient $\alpha_{\text{e}}$ (typically $0.01$) to encourage exploration; and $\alpha_{\text{c}}$ (typically $0.1$) is the weight for the consistency loss.

\begin{figure*}
    \centering
    \includegraphics[width=0.9\linewidth]{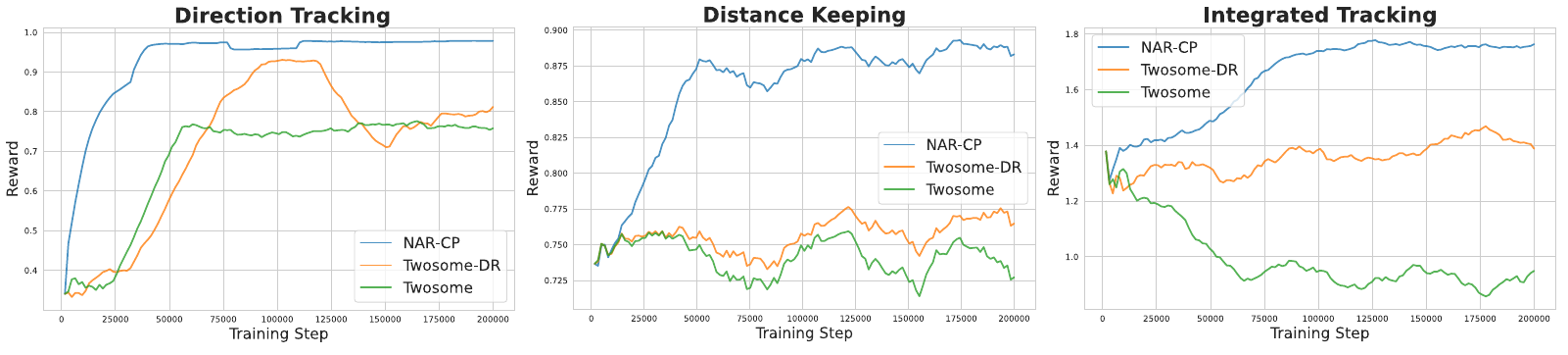}
    \caption{Figure from left to right in turn shows: The reward curve in the direction-tracking task. The reward curve in the distance-keeping task. The reward curve in the Integrated-tracking task.}
    \label{fig:placeholder3}
\end{figure*}

\begin{figure*}
    \centering
    \includegraphics[width=0.9\linewidth]{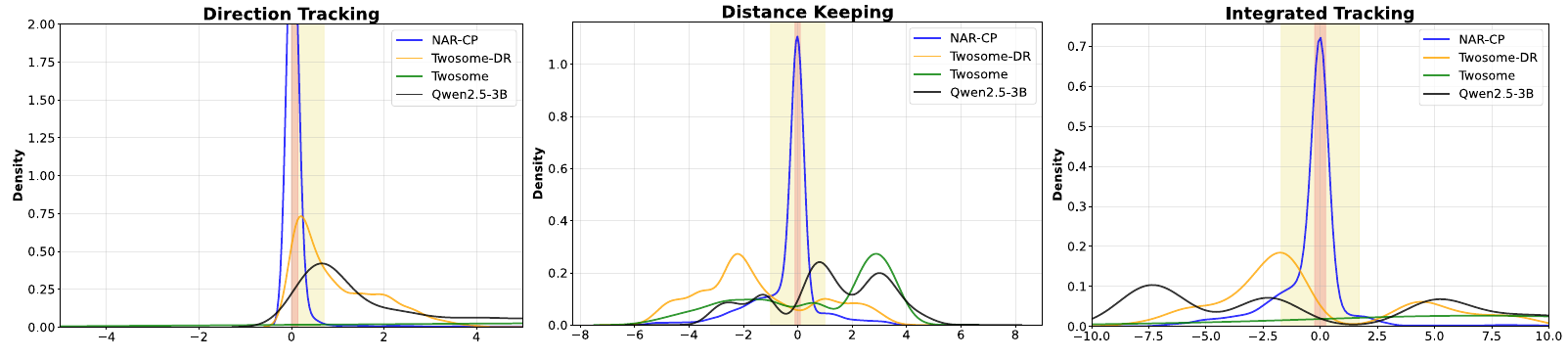}
    \caption{Error distribution characteristics: 
\textbf{Direction tracking}: $x$-axis = root sum of squared deviations, $y$-axis = density; red zone = $0$--$0.141$ (precision), yellow zone = $0$--$0.707$ (general); accuracy increases as curve approaches 0 in positive zone. 
\textbf{Distance keeping}: $x$-axis = current $-$ target distance, $y$-axis = density; red zone = $\pm 0.1$ (precision), yellow zone = $\pm 1.0$ (general); accuracy increases as curve approaches 0. 
\textbf{Integrated Tracking}: $x$-axis = direction + distance error (+ = distance > target, $-$ = distance < target), $y$-axis = density; red zone = $\pm 0.241$ (precision), yellow zone = $\pm 1.707$ (general); coordinated control increases as curve approaches 0.}
    \label{fig:placeholder4}
\end{figure*}

\section{Experiments}

\subsection{Experimental Overview}
    Experiments use Qwen2.5-3B as the base model, fine-tuned via LoRA with 1 NVIDIA RTX 4090 GPU. Four methods are compared: (1) original untuned Qwen2.5-3B, (2) TWOSOME (basic fine-tuned version), (3) TWOSOME + dense reward, and (4) our proposed method. All evaluations are conducted in the aforementioned UAV dynamic tracking scenario, where the agent operates in an open 3D dynamic airspace to stably track a target UAV, with decisions driven entirely by high-precision numerical-textual information. All experiments are built and run on NVIDIA Isaac Gym.

    By addressing the following four key research questions, we systematically validate the effectiveness, stability, and generalization of the proposed method:

\begin{enumerate}
    \item \textbf{Comparison Experiment:} Does the proposed method improve policy convergence and reduce policy bias between sub-tasks and the composite task compared with baselines?
    
    \item \textbf{Generalization Experiment:} Does the trained model generalize well to unseen tracking scenarios and varied target UAV motion patterns?
    
    \item \textbf{Decision Latency Verification:} In 10 Hz high frequency settings, does the proposed method achieve lower inference latency than large models and other baselines?
    
    \item \textbf{Ablation Study:} How can we quantify the contribution of each core module in the proposed method?
\end{enumerate}

\subsection{Experiment 1: Comparison Experiment}
\subsubsection{Experimental Setup}
This experiment aims to verify the effectiveness of the proposed method in high-frequency decision-making scenarios: by examining the training curves of direction tracking, distance keeping, and integrated Tracking to verify improved policy convergence; by evaluating success rate and stability metrics to further demonstrate this enhancement; and by utilizing policy discrepancy metrics to assess the mitigation of policy bias.

The introduction of the three tasks is as follows:
\begin{enumerate}
    \item \textbf{Direction Tracking:} The agent selects appropriate flight headings based on the target's relative position and motion to maintain tracking continuity and prevent loss of sight.
    
    \item \textbf{Distance Keeping:} Given a heading, the agent regulates its speed relative to the target to maintain a specified distance, avoiding collisions (too close) or disengagement (too far).
    
    \item \textbf{Integrated Tracking:} The agent performs direction tracking and distance keeping simultaneously, achieving stable tracking by maintaining both angular advantage and optimal distance.
\end{enumerate}

\begin{table*}[htb]
  \centering
  \resizebox{\linewidth}{!}{
  \begin{tabular}{lccccccccc}
    \toprule
    \multirow{2}{*}{Model} 
    & \multicolumn{2}{c}{Direction Tracking} 
    & \multicolumn{2}{c}{Distance Keeping} 
    & \multicolumn{3}{c}{Integrated Tracking} \\
    \cmidrule(lr){2-3} \cmidrule(lr){4-5} \cmidrule(lr){6-8}
     & General Rate \% & Precise Rate \% 
     & General Rate \% & Precise Rate \% 
     & General Rate \% & Precise Rate \%  & Policy Bias \\
    \midrule
    Qwen2.5-3B 
      & 31.62 & 4.62 
      & 25.04 & 0.04 
      & 2.83 & 0.02 & {0.05} \\
    Twosome 
      & 1.38 & 0.08 
      & 19.00 & 1.08  
      & 0.46 & 0.06 & 0.186 \\
    Twosome-DR 
      & 51.58 & 17.38  
      & 15.12 & 0.13  
      & 24.58 & 0.10 & 1.22 \\
    NAR-CP 
      & \textbf{96.79} & \textbf{91.33}  
      & \textbf{76.67} & \textbf{60.83} 
      & \textbf{82.00} & \textbf{61.46} & \textbf{0.006} \\
    \bottomrule
  \end{tabular}}
  \caption{Performance comparison across different tracking scenarios with additional policy bias metric for integrated tracking.}
  \label{tab:tracking_performance}
\end{table*}

\begin{table*}[htb]
  \centering
  \resizebox{\linewidth}{!}{
  \begin{tabular}{lccccccccc}
    \toprule
    \multirow{1}{*}{Model} 
    & \multicolumn{2}{c}{Distance Generalization} 
    & \multicolumn{2}{c}{Strategy Generalization} 
    & \multicolumn{2}{c}{Complex Composite Task} \\
    \cmidrule(lr){2-3} \cmidrule(lr){4-5} \cmidrule(lr){6-7} \cmidrule(lr){8-9}
     & General Rate \% & Precise Rate \% 
     & General Rate \% & Precise Rate \% 
     & General Rate \% & Precise Rate \%  \\
    \midrule
    Qwen2.5-3B 
      & 9.11 & 0.44
      & 3.04 & 0.42
      & 0.88 & 0.02  \\
    NAR-CP 
      & \textbf{68.00}   & \textbf{48.05} 
      & \textbf{74.50}  & \textbf{22.38}
      & \textbf{19.62} 
      & \textbf{1.08}  \\
    \bottomrule
  \end{tabular}}
  \caption{Conducting generalization experiments for verification in unseen scenario.}
  \label{tab:generalization_performance1}
\end{table*}

\subsubsection{Results and Analysis}

As shown in Fig.~3, the training reward curves indicate that NAR-CP consistently outperforms TWOSOME and TWOSOME-DR in direction tracking, distance maintenance, and the integrated task, while dense rewards only provide limited convergence improvement. Table~1 further supports these findings: (i) NAR-CP achieves substantially better performance than both baselines across all tasks, demonstrating strong decision-making in high-precision, high-frequency scenarios; (ii) the untrained Qwen2.5-3B base model even surpasses trained TWOSOME on certain tasks, exposing the latter’s limitations in strategy convergence; and (iii) dense rewards yield marginal gains—TWOSOME-DR underperforms TWOSOME in distance maintenance, and Fig.~4(b) shows that neither method overcomes the performance bottleneck. Overall, the results in Table~1 confirm that the proposed consistency strategy effectively reduces policy deviation.

\subsection{Experiment 2: Generalization Experiment}
\subsubsection{Experimental Setup}
To evaluate the generalization capability of the proposed method, we test the model on four unseen task scenarios:
\begin{itemize}
    \item \textbf{Distance Generalization:} The training stage only involved fixed distance thresholds (5 meters and 10 meters). Now, the target distance has been randomized.
    \item \textbf{Strategy Generalization:} The target UAV used a straight flight strategy in training, and now the target was switched to a circle flight strategy.
    \item \textbf{Complex Composite Task:} We construct a multi-dimensional unknown environment by simultaneously introducing dynamic distances, color randomization, and circular trajectories.
\end{itemize}

\subsubsection{Results and Analysis}
Table 2 show that leveraging the semantic capability of large models, our method can generalize to various unseen scenarios and has a certain ability to understand numerical information, achieving accurate tracking to a certain extent in distance generalization and trajectory strategy generalization tasks. However, constrained by the discrete action space design, it is difficult for the method to complete tasks accurately in more complex unseen scenarios.

\subsection{Experiment 3: Decision Latency Verification}
\subsubsection{Experimental Setup}
This experiment aims to verify the real-time performance of our method in high-frequency decision-making scenarios. We conduct a core comparison of action inference latency between the trained model and the original Qwen2.5 model. By recording the time consumption from receiving high-precision numerical text to outputting flight commands and calculating the average latency, we quantitatively evaluate the real-time decision-making capability to verify if it meets the real-time requirements.

\begin{table}[h]
    \centering
    \begin{tabularx}{\linewidth}{XXXX}
        \toprule
        Method & Direction Tracking & Distance Keeping & Integrated Tracking \\
        \midrule
        Qwen2.5-3B-infer  & 1968 ms & 1654 ms & 2489 ms \\
        NAR-CP & 114 ms & 95 ms & 144 ms \\
        \bottomrule
    \end{tabularx}
    \caption{Decision Latency}
    \label{tab:ablation1}
\end{table}

\subsubsection{Results and Analysis}
As shown in Table~3, the inference decision-making time of the large model is significantly longer than that of our method. Moreover, the inference latency is strongly correlated with the input prompt length.

\subsection{Experiment 4: Ablation Study}
This section evaluates the individual and synergistic effects of action reward normalization and consistent policy loss function via ablation experiments, validating their effectiveness in high-frequency numerical decision scenarios.

\begin{table}[h]
    \centering
    \resizebox{\linewidth}{!}{
    \begin{tabular}{ccccc}
    
        \toprule%第一道横线
        NAR   &  CP    &  General Rate  & Precise Rate   & Policy bias    \\
        \midrule%第二道横线 
            \checkmark  &\checkmark  & \textbf{82.54\%}  & \textbf{62.75\%}  & \textbf{0.008}      \\
            \checkmark  &            & {78.96\%}         & 57.67\%         & 7.174\\
                        &\checkmark  & {6.71\%}         & 0.92\%           & 0.025       \\
                        &            & 7.54\%           & 0.02\%           & 1.163       \\
        \bottomrule%第三道横线
    \end{tabular}
    }
    \caption{Ablation Test}
    \label{tab:ablation2}
\end{table}

As show in Table 4, ablation experiments demonstrate that action reward normalization and consistent policy loss function form an effective complement in addressing high-frequency decision problems: the former provides a stable convergence foundation for the model, while the latter significantly enhances the adaptability of the policy. By integrating these two techniques, the model not only achieves a rapid and stable convergence but also effectively reduces policy bias, striking an ideal performance balance between generalization control capability of composite task and precise operations in high-frequency tasks, thereby fully validating the scientific and practical merits of their collaborative design.

\section{Conclusion}
In this paper, we address the inherent limitations of LLMs in high-frequency decision tasks, specifically the difficulty in adapting to high-precision numerical state information and the misalignment of strategies in sub-combined tasks, by proposing the NAR-CP method that normalizes and shapes dense environmental feedback rewards while theoretically preserving optimal strategies, and aligns global and sub-semantic strategies through consistency loss to mitigate combined-task strategy misalignment, with drone tracking experiments demonstrating excellent performance in both sub-tasks and combined tasks along with strong generalization capabilities for unseen tasks.

However, although our method extends LLM decision-making to a high-frequency continuous environment, we set the action space is discrete, which makes it difficult to apply to more complex tasks. In the future, we hope to expand this work to the continuous action space, enabling LLMs to truly act as a decision-making intelligent agent and participate in high-frequency decision-making.
\appendix

\bibliographystyle{named}
\bibliography{ijcai26}

\end{document}